# YOLO-OB: An improved anchor-free real-time multiscale colon polyp detector in colonoscopy

Xiao Yang [a], Enmin Song [a], Guangzhi Ma [a, *], Yunfeng Zhu [a], Dongming Yu [a], Bowen Ding [a], Xianyuan Wang [b]

[a] School of Computer Science and Technology, Huazhong University of Science and Technology, Wuhan, Hubei, 430074, China
[b] Intelligent Image Analysis Department, Wuhan United Imaging Healthcare Surgical Technology Co., Ltd, Wuhan, Hubei, China

* Corresponding author.
  E-mail address: maguangzhi@hust.edu.cn
  Postal address: School of Computer Science and Technology, Huazhong University of Science and Technology, Wuhan, Hubei, 430074, China

**Abstract**

Colon cancer is expected to become the second leading cause of cancer death in the United States in 2023. Although colonoscopy is one of the most effective methods for early prevention of colon cancer, up to 30% of polyps may be missed by endoscopists, thereby increasing patients' risk of developing colon cancer. Though deep neural networks have been proven to be an effective means of enhancing the detection rate of polyps. However, the variation of polyp size brings the following problems: (1) it is difficult to design an efficient and sufficient multi-scale feature fusion structure; (2) matching polyps of different sizes with fixed-size anchor boxes is a hard challenge. These problems reduce the performance of polyp detection and also lower the model's training and detection efficiency. To address these challenges, this paper proposes a new model called YOLO-OB. Specifically, we developed a bidirectional multiscale feature fusion structure, BiSPFPN, which could enhance the feature fusion capability across different depths of a CNN. We employed the ObjectBox detection head, which used a center-based anchor-free box regression strategy that could detect polyps of different sizes on feature maps of any scale. Experiments on the public dataset SUN and the self-collected colon polyp dataset Union demonstrated that the proposed model significantly improved various performance metrics of polyp detection, especially the recall rate. Compared to the state-of-the-art results on the public dataset SUN, the proposed method achieved a 6.73% increase on recall rate from 91.5% to 98.23%. Furthermore, our YOLO-OB was able to achieve real-time polyp detection at a speed of 39 frames per second using a RTX3090 graphics card. The implementation of this paper can be found here: https://github.com/seanyan62/YOLO-OB.

Keywords: Deep learning; Colonoscopy; Colorectal cancer; Polyp detection; Multi-scale detection; Anchor-free; polyp dataset SUN

## 1. Introduction

Colon polyps are abnormal growths developing on the lining of colon or rectum. Most polyps are benign, but if untreated some of them can develop into colon cancer, causing the second most common cancer death in the United States [1]. According to the estimation of American Cancer Society, in 2023, there may be 52,550 deaths of colorectal cancer, posing a threat second only to lung [2]. Therefore, early detection and removal of colon polyps are indispensable for preventing the development of colon cancer.

Colonoscopy is presently considered the most reliable diagnostic method for detecting and removing polyps [3, 4]. However, the diagnostic accuracy is heavily dependent on the proficiency and experience of the medical practitioner. Studies have shown that manual inspection can miss up to 30% of polyps [5, 6]. This indicates there is an urgent need for effective prevention strategies to combat the increasing incidence and mortality rate of colorectal cancer.

In recent years, there is a surge of interest in using computer-aided detection (CAD) systems to enhance the precision and consistency of polyp detection. These systems utilize advanced object detection algorithms to analyze the colonoscopy images and automatically identify the polyps [7]. Compared to the examination of medical practitioner without intelligent algorithm, CAD has demonstrated a significant improvement on colon polyp detection rate [3, 8].

As a branch of object detection, the polyp detection task can be divided into two phases: the era of traditional object detection algorithms and the era of deep learning-based detection algorithms [9, 10]. Traditional object detection algorithms relied mainly on manually designed feature extraction. Texture, color wavelet, histogram of oriented gradients (HOG), and local binary pattern (LBP) are commonly used as manually designed features [11, 12]. Although these algorithms can achieve high accuracy on carefully selected datasets, requiring researchers to meticulously design features to detect polyps. Their clinical practicality is limited by slow processing time and growing volume of video data [13-15]. Due to the complexity of image data, it is a challenge to manually design advanced semantic features, where it is precisely the advantage of the Convolutional Neural Network (CNN).

A frequently encountered limitation in early artificial intelligence (AI) research was the availability of hardware. In recent years, the emergence of a new generation of modern GPUs has led to innovation in AI. Modern algorithms utilize CNNs and deep learning for CAD systems of polyps [7]. After training, algorithms can automatically extract features to identify polyp and non-polyp areas in large data sets without excessive manual design.

Most existing advanced algorithms for detecting polyps employ CNNs and have the potential to assist clinicians in real-time detection during colonoscopy. This could significantly enhance the sensitivity of polyp detection, leading to improved patient outcomes. Studies have demonstrated that even a marginal increase of 1% in the polyp detection recall rate can reduce the risk of colon cancer by 3% [16]. Despite considerable progress, current state-of-the-art polyp detection algorithms still exhibit inadequate recall rates. For instance, the reported highest recall rate on the SUN, a large dataset that is widely used for polyp detection, is 91.5% [17]. However, this may not be sufficient for disease detection purposes. High missed detection rates of polyps can lead to an increased risk of developing colon cancer. Accordingly, this article aims to enhance the recall rate of intelligent polyp detection algorithm.

Multi-scale detection is critical in the field of CNN-based object detection, which is also present in polyp detection tasks [18-20] as the colon polyps appear in various size, as shown in Fig.

1. During colonoscopy, the movement of the camera exacerbates this phenomenon, making it difficult for CNN models to extract effective features and match detection boxes. Studies show that the polyp detecting ability decreases as the size decreases [21]. Thus, one of the major challenges for CNN-based polyp detection is the multi-scale problem [22]. Considering the importance of the missed detection problem of polyps, we propose a high-accuracy real-time anchor-free polyp detection model named YOLO-OB. Our model adopts the basic YOLO series framework, and put forward several improvements for multi-scale detection. In general, our contributions are summarized as follows:

1. We designed a feature fusion structure, named BiSPFPN, which fuses three distinct depth feature maps (downsampled by 8, 16, and 32 times from the input image) derived from the backbone network. This approach enables efficient and comprehensive fusion of convolutional features with different receptive fields.

2. It is the first time to introduce the anchor-free ObjectBox detection into polyp detection. ObjectBox treats all polyps equally at different feature levels, regardless of the size or shape of polyps in the original image. Our algorithm is suitable for both large and small polyps and overcomes the positive sample reduction problem caused by threshold constraints when calculating losses in anchor-based methods.

3. We propose a new model YOLO-OB and conduct sufficient experiments on the public dataset SUN and the self-collected dataset Union. The recall rate of YOLO-OB greatly surpasses those of state-of-the-art models.

On the SUN dataset, the YOLO-OB achieves a precision of 98.37%, a recall of 98.23%, and a mAP of 98.19%. Compared with the state-of-the-art models on SUN, the recall rate of YOLO-OB increases by 6.73%, from 91.50% to 98.23%. On the self-collected dataset Union, it can also achieve a high precision of 98.64, a recall of 99.52, and a mAP of 99.41. Our model can accomplish polyp detection with an inference speed of 39 frames per second (FPS) on the RTX3090 graphics card. Even on the lower performance GTX1080Ti graphics card, it still maintains an inference speed of 29FPS. Thus, it totally satisfies the requirement of real-time inference in clinical scenarios.

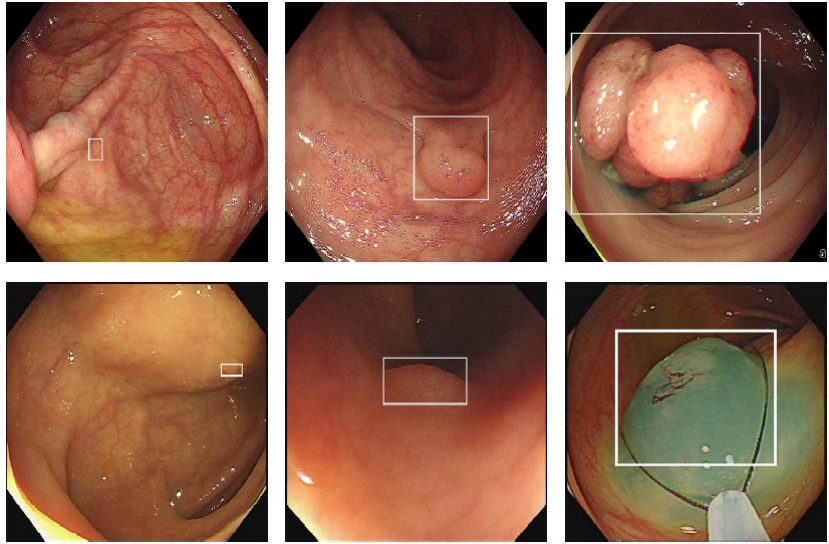

Fig. 1. The first line shows colon polyp samples from the public dataset SUN, while the second line displays our self-collected dataset Union. The white solid box is the ground truth for polyp detection task. There is a large variation in the size of colon polyp tissues. Small-sized polyps contain less information, making it difficult for models to extract efficient features.

## 2. Related work

The application of machine learning algorithms for polyp detection tasks commenced in the early 2000s [9]. Early intelligent algorithms were manually created and designed to identify polyps by detecting particular features chosen by researchers, such as texture, shape, or color information [11, 12]. Although these algorithms have usually shown high accuracy on carefully selected datasets, their applicability in the real world is limited due to the slow processing time and the difficulty in adapting to large datasets. Karkanis et al. [13] developed an algorithm that utilizes color wavelet covariance to detect textural changes which may indicate the presence of adenomatous tissue. The major limitation of this algorithm was its slow inference speed, which persisted in its updated versions [14, 15]. In 2015, Wang et al. [23] developed a traditional detection algorithm that utilizes edge detection technology to achieve polyp detection in near real-time. The algorithm attained a recall rate of 97.7% across 53 videos. It is worth noting that the evaluation of the recall rate was based on individual instances of polyps rather than each image. Although the algorithm achieved a near real-time inference speed of up to 10 frames per second, it might not be adequate for an actual real-time CAD system.

In recent years, the released datasets of colon polyps have demonstrated a diverse range of characteristics, such as size, color, and texture [24]. Traditional machine learning-based methods have struggled to meet the requirements of clinical CAD systems for both accuracy and speed [25]. In the MICCAI 2015 Polyp Detection Competition, CNN-based methods surpassed manually designed feature algorithms in terms of performance and speed [26]. Nowadays, with large polyp datasets and powerful computational hardware available, CNN-based algorithms have become the mainstream approach to solve the task of polyp detection [27].

In 2020, Misawa et al. released the largest publicly available dataset of colon polyps, named SUN [17]. The dataset consisted of 49136 images obtained from colonoscopy cases that were reviewed and annotated by both research assistants and expert endoscopists. They developed an AI-assisted polyp detection system called EndoBRAIN-EYE, which uses the YOLOv3 deep learning model architecture with modifications to the input image size. According to the official website of the SUN dataset, the latest model for polyp detection (July 2022) achieved a precision of 99.6% and a recall of 91.5%. As a task of disease detection, we aim to further improve the recall rate to reduce the risk of colorectal cancer for patients by reducing the missed polyp rate. Based on our analysis of the SUN dataset, we have discovered that the problem of multiscale detection, which is a challenging task in the field of object detection, still exists in the context of polyp detection.

There are three main approaches that address the aforementioned challenge of multiscale detection: region-proposal detection, multiresolution detection and multireference detection [28]. In the two-stage algorithm, the region-proposal stage can improve the detection accuracy, but introduces too much computation. Xi Mo et al. [29] fine-tuned the typical two-stage algorithm, faster-RCNN [30], and obtained highly competitive performance on dataset CVC-ClinicDB compared to the state-of-the-art approaches. However, the inference speed of the faster-RCNN is at most 17 frames per second (FPS). This makes it difficult to meet the real-time detection requirement for CAD systems. Multiresolution and multireference detection are the two fundamental building blocks in state-of-the-art object detection systems [10].

2.2 Feature pyramid in polyp detection

Multiresolution technology detects objects of different scales at different network layers. Before the advent of FPN technology, in polyp detection algorithms, detection heads were typically placed in the deepest layer of the network. Although the deepest features possess richer semantic information, but they are not conducive to detecting small polyps due to the lack of spatial and texture information. Lin et al. [31] proposed a network architecture known as FPN with top-down skip connections, which can easily fuse deep semantic features into shallow feature maps and make predictions on each layer simultaneously.

Later, FPN was integrated into the YOLOv3 algorithm and used to design polyp detection models [17, 32]. However, in the FPN network, the low-level features are difficult to propagate to construct the high-level features. Specifically, the shallow features of polyps must traverse the entire backbone network to affect the deep layers, where there might be a significant loss of low-level information. This problem also exists in natural image object detection tasks. Liu et al. [33] designed the PANet for natural image dataset MS-COCO [34] to address this problem. They added an additional shallow-to-deep path to transmit shallow information to deep layers. Subsequently, PANet was integrated into YOLOv4 and achieved fairly competitive results in polyp detection tasks [35, 36]. During the feature fusion process, PANet increases computational demand by introducing an additional pathway from the shallow to the deep layers. Moreover, it simply adds features from various layers without considering their respective contributions to the final result.

In order to enhance the fusion performance of FPN, Tan et al. [37] simplified the network topology of PANet and proposed a novel Bidirectional Feature Pyramid Network (BiFPN). By assigning different weights to the features across different layers, the performance was further enhanced. It achieved state-of-the-art accuracy with much fewer parameters and FLOPs than previous object detection models on dataset MS-COCO. The Feature Pyramid technology can detect objects of different scales at different layers, effectively handling the problem of multi-scale polyp targets. This technology is now an essential component of the state-of-the-art polyp detection systems [36, 38-40]. In a word, efficiently and adequately fusing features at different layers is a key problem for CNN-based polyp detection algorithms.

2.2 Bounding box regression in polyp detection

The main idea of multireference detection is to first define a set of references (also known as anchor boxes) for an image. Then, the detection box is predicted based on these references[41, 42]. Early polyp detection algorithms use sliding windows to identify objects. In order to improve the detection accuracy, researchers need to scan all possible locations in the entire image with sliding windows of different sizes [43, 44]. It is very time-consuming and cannot meet the real-time detection needs of CAD systems. Since the Faster RCNN [45] proposed in 2015, end-to-end methods have been commonly used to predict bounding boxes from CNN features. At present, due to the tremendous success of YOLO series algorithms, anchor-based one-stage detection has become the mainstream solution for polyp detection tasks [32, 38, 46].

Anchor-based object detection is commonly carried out by categorizing and regressing on candidate regions, which are called anchor boxes generated through sliding windows. Due to the considerable variability in the scale of polyp targets, the size, quantity, and aspect ratios of these anchors must be meticulously designed. During the training process, anchor-based models require calculating the intersection over union (IOU) between all anchors and ground-truth boxes, which leads to significant consumption of memory and time. In recent years, researchers have turned their attention towards anchor-free solutions as they have observed that the distribution of colon polyps in a colonoscopy image is sparse [47, 48]. Instead of relying on anchor boxes, it detects objects by identifying key points. These methods use deep learning features to regress boundary boxes directly, thereby simplifying the multi-scale detection problem. Wang et al. [47] employed the anchor-free method to detect colorectal polyps. Their methodology was inspired by the CenterNet [49], which detects the center points of polyp targets, and then using the image features of the center points to obtain the polyp boundary. Their polyp detection system surpassed previous studies and achieved state-of-the-art performance in terms of accuracy and inference speed on CVC-Clinic dataset.

Regardless whether it is an anchor-based or anchor-free solution, specific thresholds are commonly used to select positive samples for training. However, it is difficult to completely cover all polyp samples on different resolution feature maps. This will miss many positive samples during the training phase, which is unfavorable for multi-scale detection and will reduce the training efficiency. Zand et al. [50] designed a new bounding box regression algorithm called ObjectBox to detect objects of different scales on the MS-COCO dataset. It can train the model with objects of all scales as positive samples without any constraints imposed by static or dynamic assignment strategies. ObjectBox redefines the regression goal of the boundary box, allowing it to learn positive samples of all scales and thus improving the training efficiency. Since the hyperparameters related to object scales were removed, it can be easily applied to new datasets without making many adjustments.

## 3. The proposed methodology

In this paper, we focus on the challenge of multi-scale polyp detection by using multi-resolution detection technique and design an anchor-free one-stage model called YOLO-OB. DarkNet53 is utilized as our feature extraction backbone network. While in the neck part, BiSPFPN is adopted to create efficient feature fusion layers. As for the detection head, ObjectBox is employed to generate the boundary boxes of polyps of different scales.

### 3.1 The overall architecture

Similar to typical YOLO series algorithms, the proposed YOLO-OB can be divided into three parts: backbone, neck, and detection head. The input polyp image for YOLO-OB has been enhanced with data augmentation. The backbone part is responsible for extracting features from the input image, the neck part is used for feature fusion, and the detection head outputs the detection results. We modified the neck and detection head parts of YOLOv3 and developed a new YOLO-OB model to enhance polyp detection performance. Fig. 2 illustrates the overall structure of our model.

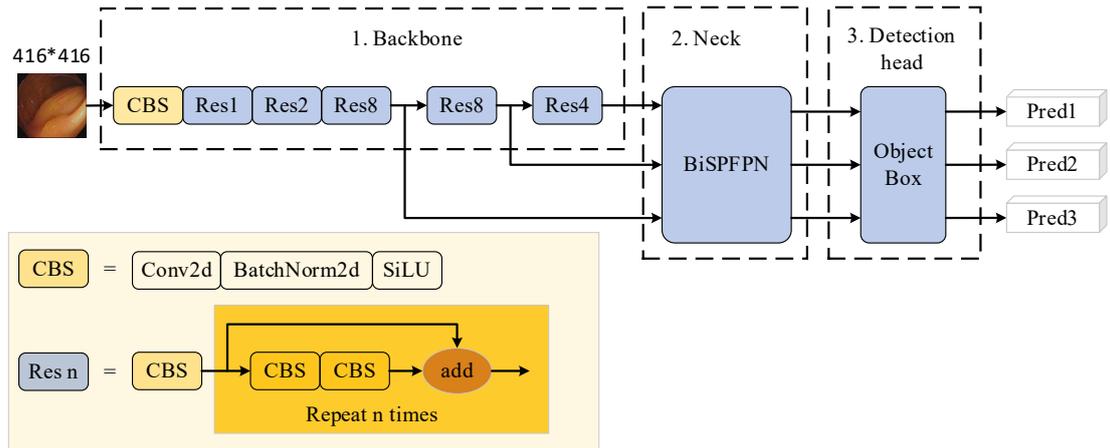

Fig. 2. The overall structure of the proposed YOLO-OB model.

As shown in Fig. 2, the augmented polyp image is resized to 416*416 pixels as the input. The YOLO-OB employs DarkNet53 as the backbone, in which the fully connected layer is removed. In the neck part, an efficient feature fusion structure BiSPFPN is designed to fully fuse the backbone encoded features of three different sizes, as discussed in Section 3.2. In the detection head part, the locations of polyps are predicted by using an anchor-free detection algorithm, which is discussed in detail in Section 3.3.

Table 1 presents the detailed information for each backbone layer. The backbone's basic convolutional unit CBS is made of Conv+BN+SiLU, which uses the BatchNorm2d to normalize convolutional features and the SiLU [51] activation function to enhance the model's non-linear fitting ability. The backbone employs numerous residual structures (Resn), which serve two purposes within two steps:

(1) It uses a CBS module with a stride of 2 instead of pooling operations to achieve feature map down-sampling, avoiding information loss caused by manually designed pooling layers.

(2) It performs channel shrinking by using a 1*1 convolution kernel, followed by channel expansion by using a 3*3 kernel.

These two steps and the residual unit are repeated n times for each Resn, allowing an increase of the model's depth while effectively preventing an excessive increase of the number of the model's parameters.

In the backbone part, feature maps down-sampled by 8, 16, and 32 times are used for detection of different sizes of polyps. Table 1 displays the matrix dimensions for each down-sampled map sized as (52*52*256), (26*26*512), and (13*13*1024), and they are correspondingly marked as $F_{52}$, $F_{26}$ and $F_{13}$. The larger feature map $F_{52}$ extracts the low-level texture features and has a smaller receptive field, which is used to recognize small-sized polyps. On the other hand, the smaller feature map $F_{13}$ captures high-level semantic features and has a larger receptive field. Therefore, it is utilized to detect large-sized polyps.

Table 1

Detailed backbone parameters of our model. The numbers in the first column represent the residual modules repeated n times.

|  | Layer | Filter size | Stride | Output | Remarks |
|---|---|---|---|---|---|
|  | Convolutional | 3*3*32 | 1 | 416*416*32 |  |

| | Convolutional | 3*3*64 | 2 | 208*208*64 | |
|---|---|---|---|---|---|
| | Convolutional | 1*1*32 | 1 | | |
| 1* | Convolutional | 3*3*64 | 1 | | |
| | Residual | | | 208*208*64 | |
| | Convolutional | 3*3*128 | 2 | 104*104*128 | |
| | Convolutional | 1*1*64 | 1 | | |
| 2* | Convolutional | 3*3*128 | 1 | | |
| | Residual | | | 104*104*128 | |
| | Convolutional | 3*3*256 | 2 | 52*52*256 | |
| | Convolutional | 1*1*128 | 1 | | |
| 8* | Convolutional | 3*3*256 | 1 | | |
| | Residual | | | 52*52*256 | $F_{52}$ |
| | Convolutional | 3*3*512 | 2 | 26*26*512 | |
| | Convolutional | 1*1*256 | 1 | | |
| 8* | Convolutional | 3*3*512 | 1 | | |
| | Residual | | | 26*26*512 | $F_{26}$ |
| | Convolutional | 3*3*1024 | 2 | 13*13*1024 | |
| | Convolutional | 1*1*512 | 1 | | |
| 4* | Convolutional | 3*3*1024 | 1 | | |
| | Residual | | | 13*13*1024 | $F_{13}$ |

### 3.2 The neck

At the neck part, fusing feature maps of different depths can enhance the ability of detecting multi-scale targets, especially the small polyps. We followed the idea of feature pyramid network and designed a novel multi-scale feature fusion structure named BiSPFPN. Its network architecture is shown in Fig. 3. On the left side of the BiSPFPN, $F_{13}, F_{26}, F_{52}$ are the feature maps at different depths (refer to the "Remarks" column in Table 1); $F'_{13}, F'_{26}, F'_{52}$ on the right side are new feature maps after feature fusion, which are used as the input of detection head module.

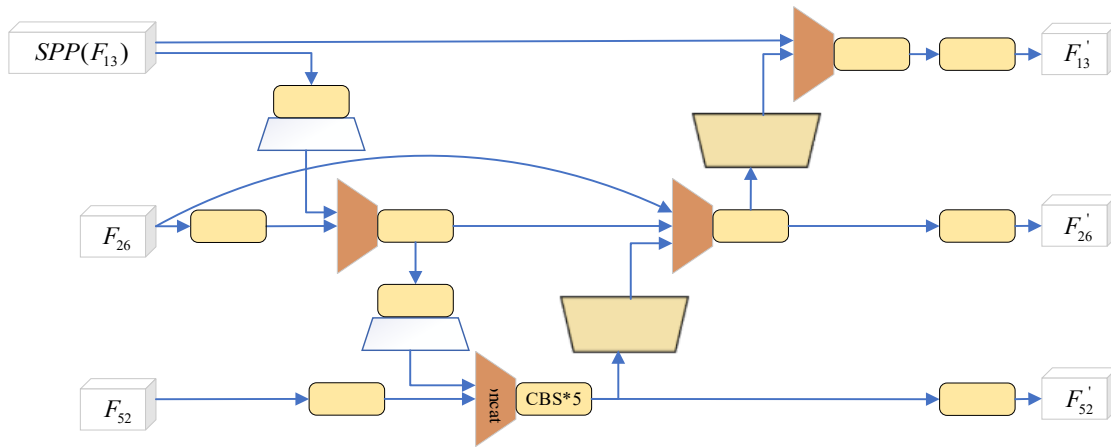

Fig. 3. The BiSPFPN architecture serves as the neck component in the YOLO-OB model. The CBS block consists of convolution, batch normalization, and SiLU activation. The "concat" operation concatenates feature maps along the channel dimension. The "Upsample" operation uses nearest-neighbor interpolation to increase the size of the feature map. The "Downsample" operation reduces

the size of feature maps by using a convolutional kernel with a size of 3x3.

We applied a SPP module on $F_{13}$, as illustrated in Fig. 4, achieves the fusion of local and global features on the feature map at the deepest convolutional layer of the backbone. In the left of the SPP, three CBS blocks employ the bottleneck architecture. The CBS consists of convolutional kernels with sizes of (1*1), (3*3) and (1*1) to decrease, augment, and decrease the channel dimension, producing a feature matrix of dimensions (13*13*512). Then, the max-pool layers with different kernel sizes {5,9,13} are used to filter the local features of different scales, while padding pixels {2,4,6} are used correspondingly to maintain the output size of the max-pool layer at (13*13). We concatenate these three local features with the primary features and generate a feature matrix with dimensions of (13*13*2048). Then, two CBS modules are used to compress the feature matrix. The final feature map output by the SPP module still has the dimensions of (13*13*1024).

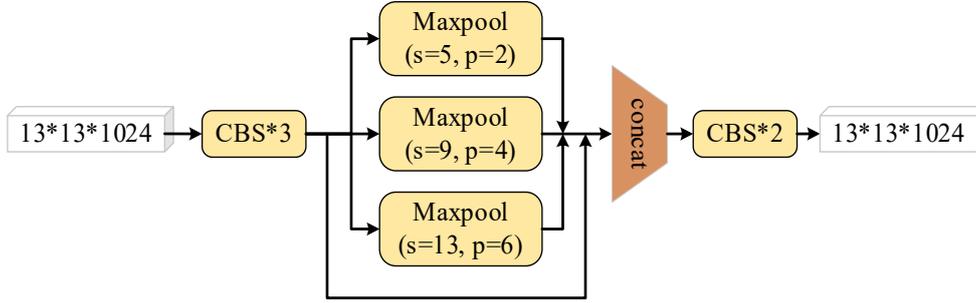

Fig. 4. SPP module structure used in YOLO-OB. The CBS block consists of convolution, batch normalization, and SiLU activation.

In the typical FPN network, there is a deep-to-shallow path, which can effectively transfers the high-level features to the low-level, whereas the low-level features can only transfer to the high-level through entire backbone. For example, in the FPN, the low-level feature $F_{52}$ involved in constructing $F_{13}$ must pass through 26 convolution layers. So, we introduce a shallow-to-deep path, making it easier for the low-level information to be transmitted to the high-level. Now, the same process only needs to pass through 13 convolution layers. Observing Fig. 3, the left half can be interpreted as a "deep-to-shallow" path, while the right half can be interpreted as a "shallow-to-deep" path.

Considering that the "concat" operation is meaningless when there is only one input, we removed nodes for feature fusion that had only one input. This strategy can simplify the bidirectional feature fusion network and reduce computation cost. Inspired by U-Net [52], we added a skip connection at level $F_{26}$, which allows more comprehensive feature fusion without introducing additional computation. Ultimately, we obtained a U-shaped multi-scale feature fusion structure, called BiSPFPN. It achieves rapidly bidirectional cross-scale feature fusion.

As shown in Fig. 3, we can regard this bidirectional feature pyramid network as an independent structure (BiSPFPN layer). By stacking BiSPFPN layers, more adequate feature fusion results can be obtained. However, we observed that additional BiSPFPN layers do not significantly improve polyp recognition accuracy, but increase the computation cost, and prolong the inference time. Thus, we utilize only one single BiSPFPN layer to fuse features from three feature maps of different scales.

3.3 The detection head

The current anchor-based algorithms choose positive samples and compute losses using pre-defined IoU threshold on feature maps of various sizes. However, the anchors' scales corresponding to feature maps of different sizes are different. Generically, larger objects are typically matched with smaller feature maps, due to the latter having larger receptive fields. Therefore, larger objects are intended to be identified as negative samples on the bigger feature maps.

In the anchor-based algorithm, the predefined IoU threshold strategy filters out some training samples, preventing the full utilization of features from three scales. Therefore, we implemented the label assignment module known as ObjectBox. The output of the neck part includes three feature maps in sizes (13*13*1024), (26*26*512), and (52*52*256) respectively. Polyps are detected independently by our model on each of these feature maps with different scales.

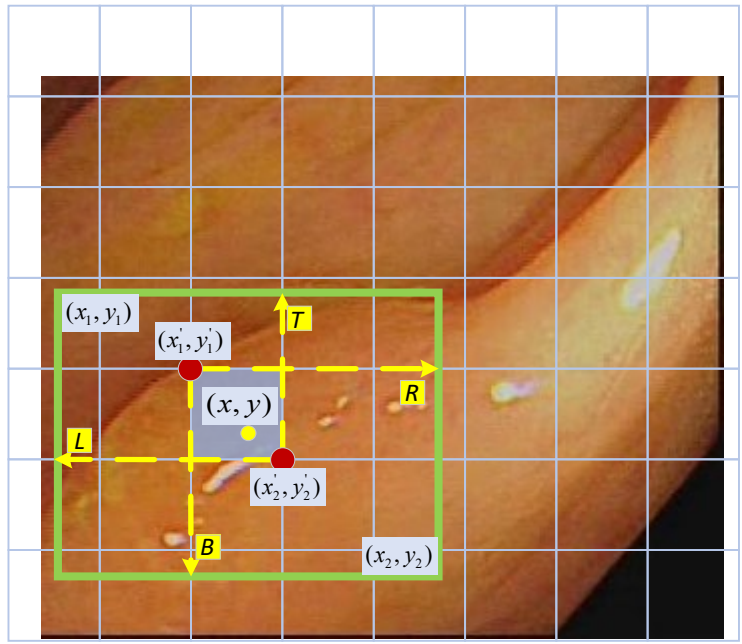

Fig. 5. Bounding box regression strategy. The background is the mapping of real polyp sample on a feature map of scale i. The green solid box represents the ground-truth box, and the coordinates of its upper-left and lower-right are $(x_1, y_1)$ and $(x_2, y_2)$. The center point of the ground-truth box is marked as $(x, y)$. The blue cell on the feature map correspond to the cell where the center point $(x, y)$ is located. Its upper-left and lower-right coordinates are $(x_1', y_1')$ and $(x_2', y_2')$. L, B, R, T are the regression targets.

Fig. 5 illustrates the specific details of this box regression strategy. For each feature map, we only use the cell containing the target center (blue cell in Fig. 5) to regress the bounding box, while the rest of the feature cells are marked as negative samples. According to this method, for each polyp target, regardless of its scale, three positive sample points can be obtained on three different feature maps. Thus, every feature map can have the ability to predict objects of different scales, this improves the model's ability to detect polyps of different sizes. In detail, according to Fig. 5, $L = x_2' - x_1, R = x_2 - x_1', T = y_2' - y_1, B = y_2 - y_1'$ are the regression targets of the neural network, and their calculation methods are as follows:

$$\begin{cases} L^{(i)} = \left( \left\lfloor \dfrac{x}{s_i} \right\rfloor + 1 \right) - \left( \dfrac{x_1^{(i)}}{s_i} \right) \\ T^{(i)} = \left( \left\lfloor \dfrac{y}{s_i} \right\rfloor + 1 \right) - \left( \dfrac{y_1^{(i)}}{s_i} \right) \\ R^{(i)} = \left( \dfrac{x_2^{(i)}}{s_i} \right) - \left\lfloor \dfrac{x}{s_i} \right\rfloor \\ B^{(i)} = \left( \dfrac{y_2^{(i)}}{s_i} \right) - \left\lfloor \dfrac{y}{s_i} \right\rfloor \end{cases} \quad (1)$$

where $\left( L^{(i)}, T^{(i)}, R^{(i)}, B^{(i)} \right)$ denote the regression targets of the bounding box based on the feature map of scale i. $\left( \left\lfloor \dfrac{x}{s_i} \right\rfloor, \left\lfloor \dfrac{y}{s_i} \right\rfloor \right)$ and $\left( \left\lfloor \dfrac{x}{s_i} \right\rfloor + 1, \left\lfloor \dfrac{y}{s_i} \right\rfloor + 1 \right)$ are the values of $(x_1', y_1')$ and $(x_2', y_2')$. $s_i = \{32,16,8\}$ represents the strides between the two cells on the three feature maps, namely $F_{13}, F_{26}, F_{52}$. Smaller feature maps such as $F_{13}, F_{26}$ have larger strides such as 32,16 respectively. The model predicts the following four regression targets:

$$\begin{cases} L^{(i)*} = sigmoid(p_1) * 2^i \\ T^{(i)*} = sigmoid(p_2) * 2^i \\ R^{(i)*} = sigmoid(p_3) * 2^i \\ B^{(i)*} = sigmoid(p_4) * 2^i \end{cases} \quad (2)$$

The variables $(p_1, p_2, p_3, p_4)$ represent the outputs of the model which can be used to calculate distances $(L, T, R, B)$. The predicted $p_1, p_2, p_3$ and $p_4$ are normalized between 0 and 1 by using the sigmoid function. To adapt to different feature map sizes, a scaling factor of $2^i$ (where i = {4, 2, 1} for feature maps $F_{13}, F_{26}, F_{52}$) is multiplied to the results of the sigmoid function.

The model generates output vector of size 6 for each position in the above three feature maps, consisting of the four distance values, a confidence score for the bounding box, and a category label. As the model only detects the presence of the polyp target, without further subdivision of its specific type, therefore, the length of the category label is 1.

### 3.4 The loss function

Ignoring the batch size parameter, the detection head generates three matrices with dimensions (52*52*6), (26*26*6) and (13*13*6), which represent the predicted results for feature maps of different scales. In contrast to conventional anchor-based approaches, we use all predicted boxes as positive samples to calculate loss, without applying any constraints to filter out some predicted boxes. Our model's overall loss function is as follows:

$$loss_{box} = \sum_{i=0}^{S^2}\left(1 - SDIoU\left(b_p, b_g\right)\right)$$

$$loss_{obj} = \sum_{i=0}^{S^2} BCE\left(c_p, o\right) \qquad (3)$$

$$loss_{total} = \lambda_1 \cdot loss_{box} + \lambda_2 \cdot loss_{obj}$$

The variable S represents the sizes of feature maps, where the values for $S^2$ are $\{13*13, 26*26, 52*52\}$. $loss_{box}, loss_{obj}$ refer to the location error and the confidence error of the predicted boxes respectively. The hyper-parameters $\{\lambda_1, \lambda_2\}$ are utilized to regulate the respective weights. In this work, we set $\lambda_1, \lambda_2$ respectively as 0.05 and 1.

The $loss_{obj}$ is calculated by using binary cross entropy, where $c_p$ is the predicted confidence value, and $o \in [0,1]$ represents the IoU between the predicted box and the ground-truth box.

With respect to $loss_{box}$, we adopt the scale-invariant SDIoU[50] as the similarity measurement to minimize the error between the predicted box $b_p = \left(L^{(i)*}, T^{(i)*}, R^{(i)*}, B^{(i)*}\right)$ and the ground-truth box $b_g = \left(L^{(i)}, T^{(i)}, R^{(i)}, B^{(i)}\right)$. Fig. 6 shows the detailed calculation method of SDIoU.

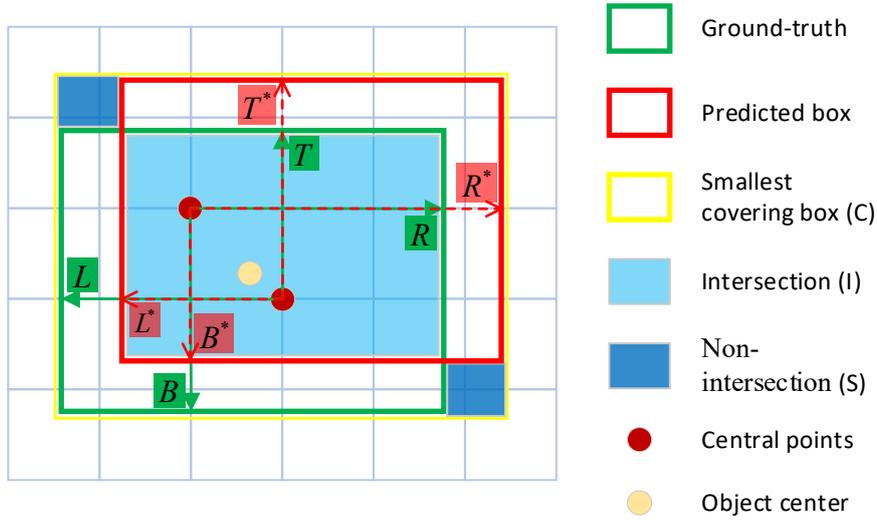

Fig. 6. The SDIoU is composed of three parts: the intersection area (I) between the predicted box and the ground-truth box, the smallest covering box (C) that contains both boxes, and the non-intersection area (S) within box C that is not covered by both boxes.

These three areas used for calculating SDIoU can be calculated using the predicted values $\left(L^*, T^*, R^*, B^*\right)$, and the ground-truth box values $\left(L, T, R, B\right)$. Specifically, the area S can be calculated using the sum of difference squares between the corresponding distance values of all four directions:

$$S = \left(L^* - L\right)^2 + \left(T^* - T\right)^2 + \left(R^* - R\right)^2 + \left(B^* - B\right)^2 \tag{4}$$

The intersection area is $I = w_I^2 + h_I^2$, where $w_I$ and $h_I$ are the width and height of the rectangle I. And their specific detailed calculation is as follows:

$$\begin{aligned} w_I &= \min(L, L^*) + \min(R, R^*) - 1 \\ h_I &= \min(T, T^*) + \min(B, B^*) - 1 \end{aligned} \tag{5}$$

The minimum box that covers both the predicted box and the ground-truth box is $C = w_C^2 + h_C^2$, where $w_C$ and $h_C$ are the width and height of box C. The calculation of $w_C$ and $h_C$ are as followes:

$$\begin{aligned} w_C &= \max(L, L^*) + \max(R, R^*) - 1 \\ h_C &= \max(T, T^*) + \max(B, B^*) - 1 \end{aligned} \tag{6}$$

The calculation of SDIoU, which measures the degree of overlap between the predicted detection box and the ground-truth box, can be calculated according to the following formula:

$$SDIoU = \frac{(I - S)}{C} \tag{7}$$

The loss function of the model's bounding box is defined as $loss_{box} = 1 - SDIoU$. By minimizing this value, the model obtains better predictions of the four distance values which match the actual distances of ground truth on feature maps of varying sizes.

**4. Experiments and discussion**

Section 4.1 describes the datasets and the augmentation strategy, Section 4.2 depicts the training strategy, software, and hardware platform used for model training, and hyperparameter settings. Section 4.3 introduces several evaluation metrics used for the polyp object detection task. Section 4.4 presents detailed ablation experiments and analysis. In Section 4.5, we compare our polyp detection model with the state-of-the-art polyp detection algorithms. Section 4.6 discusses the current limitations of the model and potential directions for future research.

4.1 Dataset specification and augmentation

This study utilized the SUN dataset, a large open dataset consisting of 49,136 images of 100 polyp samples, released by Showa University and Nagoya University in 2020 [17]. Additionally, we have also collected a private dataset called Union. We collaborated with Wuhan United Imaging Healthcare Surgical Technology Co., Ltd to collect 25 colonoscopy videos in order to expand the SUN dataset and assess the detection performance on images acquired by different devices. These videos were collected from 25 patients in different hospitals from years 2021 to 2022. The resolution of the videos varies from 720*576 to 1920*1080. To protect the patient's privacy, we have cropped the metadata area in the video and only kept the colonoscopy image area. We decomposed these videos into frames and used the open-source software Labelme to annotate bounding boxes on the frames those contained polyps, resulting in 53,188 polyp images. Professional endoscopists revised the annotations to ensure the labeling quality. Unfortunately, we are currently unable to publicly release the Union dataset. In the end, we used a total of 102,324 polyp images for the experiments, roughly in a 1:1 ratio from the SUN public dataset and our own dataset Union. According to our different experimental purposes, we correspondently carried out dataset splits for training and testing sets.

As the YOLO-OB model requires a 416*416 pixel resolution input image, but the original images may have different aspect ratios, we used center alignment to fill the image resolution into 1:1. To address the issue of small sized polyps, we utilized mosaic data augmentation to process the polyp images. As shown in Fig. 7, mosaic data augmentation arranges four randomly cropped and scaled polyp images into one image. However, we noticed that many small polyps presented at the edges of polyp images. In order to retain these samples, we abandoned the step of random cropping and only merged four images after scaling randomly. This data augmentation method constructed many small polyp samples, helping the model learn to recognize smaller polyp targets.

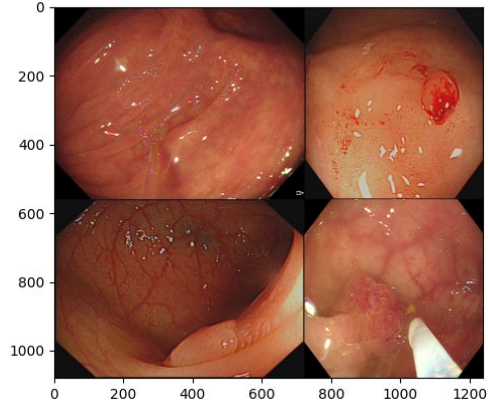

Fig. 7. Example of mosaic data augmentation: randomly select 4 images, perform random scaling on them, and then arrange them randomly into one image as input. The corresponding box labels are also properly processed.

4.2 Training strategy

We built the model using the open-source PyTorch framework and trained it with a RTX3090 graphics card. The model takes images in size of 416*416*3 as input and is trained by using the SGD optimizer with the momentum of 0.9 and the decay of 0.0005. We set the batch size to 16 and trained the model for 150 epochs. To make the model more stable in later training stage, we used the exponential decay strategy to dynamically adjust the learning rate:

$$lr = lr_{base} \times 0.99 \wedge (300 \times \frac{iters_{done}}{iters_{max}}) \quad (8)$$

where $lr$ is the SGD learning rate used to adjust the model's weights at each iteration; $lr_{base}$ is the initial learning rate, which is initially set to 0.01; $iters_{done}$ denotes the number of iterations completed by the gradient descent algorithm, while $iters_{max}$ is the total number of iterations.

4.3 Evaluation metrics

Selecting appropriate performance metrics is crucial for evaluating and comparing models,

depending on the type of task to be carried out. For the colorectal polyp detection task, the main objectives are classification and localization. To present the result of classification task, it is important and suitable to use the confusion matrix.

Table 2
Example of confusion matrix for polyp detection result. TP stands for true positive and represents that the model correctly labels a polyp prediction box. FP stands for false positive and means that the model falsely labels a prediction box in a location where there is no polyp. FN stands for false negative and means that the model does not label a detection box on a true polyp, indicating that the model missed this polyp target. As for TN, our dataset does not contain images which do not include polyps, so the value of TN is 0.

|                  | Predicted polyp | Predicted non-poly |
|------------------|-----------------|--------------------|
| Actual polyp     | TP              | FN                 |
| Actual non-poly  | FP              | TN                 |

In order to compare fairly and objectively with the current advanced methods, we used the evaluation criteria of the MICCAI Sub-Competition on Automatic Polyp Detection in Colonoscopy to evaluate the performance of polyp detection. The Precision indicator represents the proportion of positions where the model correctly identified the presence of a polyp among all locations where the model predicated as a polyp:

$$Precision = \frac{TP}{TP+FP} \qquad (9)$$

The Recall indicates the proportion of true polyp samples detected by the model among all true polyp samples:

$$Recall = Sensitivity = \frac{TP}{TP+FN} \qquad (10)$$

For classification tasks, the Precision and Recall often have an inherent trade-off. We use the F-score to comprehensively evaluate the classification performance of the model, which is the harmonic mean of Precision and Recall. The formula we used for F1 score is as follows:

$$F1 = \frac{2*Precision*Recall}{Precision+Recall} \qquad (11)$$

As it is not only a classification task, but also a polyp localization task, setting different confidences and IOU thresholds will affect the results of the confusion matrix, and thus change the results of Precision and Recall. Therefore, we introduced the most commonly used evaluation metric in the object detection field, mAP, to comprehensively evaluate the performance of the polyp detection model. The calculation of mAP is shown in Formula (12), in which K is the number of the target categories. K is set to 1 because the model only detects polyp as the single target. Eventually, mAP (or the $AP_{polyp}$) is the area under the Precision-Recall curve obtained at different confidence thresholds.

$$mAP = \sum_{i=1}^{K} AP_i = AP_{polyp} \qquad (12)$$

Another performance metric to be considered is the inference speed of the model. To deploy the polyp detection model in a clinical setting, the detection model must ensure real-time inference. We use FPS to measure the inference speed of the model, which represents the number of polyp images that the model can process in one second.

### 4.4 Ablation study

In order to verify the effectiveness of the proposed approaches, we have conducted careful ablation analysis. The performance of detection task is shown in Table 3, and the comparison of computational efficiency is presented in Table 4. Each row of these two tables corresponds to the same ablation approach for each other. To ensure the objectivity and stability for the results, we randomly shuffled the SUN dataset and carried out experiments for three times, each time the dataset was split with a ratio of 9:1 for training and testing. The final results were presented in the form of mean ± standard deviation of the three experimental results.

The detection performance of the original YOLOv3 model on the SUN dataset is shown as experiment 1 in Table 3, achieving a high precision (98.78%) but a disappointing recall (84.21%). Low recall rate can impact the judgment of colonoscopy physicians, resulting in missed polyps and increasing the risk of patients developing colon cancer. It is unacceptable. After introducing the mosaic data augmentation, experiment 2 shows an improved recall rate compared to YOLOv3 model (84.21% -> 86.47%). In experiment 2, smaller polyp samples created with mosaic data augmentation do not increase the complexity of the model nor the computational overhead during prediction.

By comparing the results of experiment 2 and 3 in Table 3, it can be seen that the BiSPFPN structure further improved the recall rate to 89.77% while maintaining a higher precision. As described in Section 3.2, the BiSPFPN can be stacked for more adequate feature fusion. As shown in Table 3 and Table 4 for experiment No.4, we construct a model with two BiSPFPN layers, but it did not improve the performance too much but added excessive parameters (78.04M -> 105.23M) and computational costs (593.91G -> 770.79G), resulting in slower training speed and inference speed (40 FPS -> 31 FPS on RTX3090). Therefore, our proposed YOLO-OB model only utilized one BiSPFPN layer to achieve multi-scale feature fusion.

After analyzing the results of experiment 5 in Table 3, we discovered that replacing the detection head with ObjectBox significantly improved the recall rate, achieving a value of 98.23%. However, this replacement led to a slight decrease in precision by 0.41% compare to the original YOLOv3, as it dropped from 98.78% to 98.37%. Generally speaking, it significantly reduces the incidence of missed detection without significantly compromising the precision. Furthermore, this method eliminates the need of manually designed anchors, thereby allowing an easy adaptation to new data distributions by minimal parameter adjustments.

We set the batch size to 1 during validation and computed the inference speed by measuring the elapsed time for the entire test set. To evaluate the model's hardware requirements, we assessed its inference speed on both RTX 3090 and GTX 1080 Ti. The performance of YOLO-OB is presented in Table 4 with experiment 5, achieving an inference speed of 39 FPS on RTX 3090 and 29 FPS on GTX 1080 Ti.

We attempted to use a larger batch size on the RTX3090 graphic card, we noticed that YOLO-OB achieved a speed of 140 FPS when the batch size was set to 32. This finding suggests that the model's inference speed may only be limited by the slow disk I/O speed. Consequently, the YOLO-OB model can provide even faster inference in the scenario where disk I/O is unnecessary, such as clinical CAD systems.

Table 3

Ablation study for performances. All experiments were conducted on the SUN dataset with 9:1 ratio for training and testing sets. We repeated this process three times. The final result is the mean ± standard deviation of the three experimental results. Experiment 1 is the result of original YOLOv3, serving as the baseline model. Experiment 5 is our final model. All performance data are presented as percentages.

| Exp. No. | Mosaic | BiSPFPN | Object Box | precision | recall (sensitivity) | F1 | mAP |
|---|---|---|---|---|---|---|---|
| 1 | | | | **98.78±0.41** | 84.21±0.33 | 90.91±0.37 | 83.13±0.69 |
| 2 | √ | | | 98.08±0.86 | 86.47±0.45 | 91.53±0.23 | 86.17±0.60 |
| 3 | √ | √ | | 98.34±0.37 | 89.77±0.30 | 93.12±0.42 | 89.60±0.30 |
| 4 | √ | √ √ | | 98.10±0.38 | 90.22±0.26 | 93.67±0.30 | 90.02±0.27 |
| 5 | √ | √ | √ | 98.37±0.47 | **98.23±0.39** | **98.32±0.32** | **98.19±0.39** |

Table 4

Ablation study for computational cost. The batch size 16 was used to evaluate computational complexity (multi-adds), and the batch size 1 was used to evaluate inference speed. Experiment 1 is the result of original YOLOv3, and experiment 5 is our final model.

| Exp. No. | Mosaic | BiSPFPN | Object Box | parameters (M) | multi-adds (G) | speed on RTX3090 (bs=1) | speed on GTX1080Ti (bs=1) |
|---|---|---|---|---|---|---|---|
| 1 | | | | 61.52 | 522.32 | 41 | 31 |
| 2 | √ | | | 61.52 | 522.32 | 41 | 31 |
| 3 | √ | √ | | 78.04 | 593.91 | 40 | 29 |
| 4 | √ | √ √ | | 105.23 | 770.79 | 31 | 23 |
| 5 | √ | √ | √ | 78.01 | 593.68 | 39 | 29 |

We also evaluate the performance of the original YOLOv3 and our YOLO-OB on the Union dataset and their generalization ability across datasets. The results are presented in the Table 5. For the Union dataset, we divided the samples into a training set (47,869 images) and a test set (5,319 images) in a ratio of 9:1. The parameters for training the original YOLOv3 are specifically optimized for the Union dataset. Experiments 1 and 2 in Table 5 show that the performances of the original YOLOv3 and our YOLO-OB on the Union dataset are comparable to their performances on the SUN dataset. Our proposed model substantially increases the recall rate from 89.77% to 99.52%.

Experiments 3, 4, and 5 demonstrate that YOLO-OB maintains a high recall rate on the combined SUN and Union dataset. However, there is a loss of approximately 2 percentage points in precision. This could be due to the inconsistent data distribution caused by different image acquisition devices used for the two datasets. As demonstrated in experiments 6, 7, 8, and 9 in Table 5, the results indicate that both models suffer from serious omission errors when encountering "new" data. Nonetheless, the results still indicate that our model has a better recall rate than the original YOLOv3, which aligns with the primary focus of the YOLO-OB design, aiming to reduce the polyp omission error.

Table 5

Comparison of the original YOLOv3 and our YOLO-OB on the large dataset constructed from Union and SUN. All performance data are presented as percentages.

| Exp. NO. | Method | training set | test set | precision | recall (sensitivity) | F1 | mAP |
|---|---|---|---|---|---|---|---|
| 1 | YOLOv3 | Union | Union | 98.15 | 89.77 | 93.77 | 89.40 |
| 2 | YOLO-OB | Union | Union | 98.64 | 99.52 | 99.08 | 99.41 |
| 3 | YOLO-OB | SUN + Union | SUN | 96.56 | 98.25 | 97.40 | 97.29 |
| 4 | YOLO-OB | SUN + Union | Union | 96.61 | 97.50 | 97.05 | 97.20 |
| 5 | YOLO-OB | SUN + Union | SUN + Union | 96.59 | 97.82 | 97.20 | 97.17 |
| 6 | YOLOv3 | SUN | Union | 78.65 | 19.87 | 31.73 | 17.59 |
| 7 | YOLO-OB | SUN | Union | 82.49 | 35.75 | 49.88 | 34.00 |
| 8 | YOLOv3 | Union | SUN | 88.14 | 45.20 | 59.75 | 42.77 |
| 9 | YOLO-OB | Union | SUN | 95.91 | 48.25 | 64.20 | 47.68 |

4.5 Comparison with state-of-the-art methods

In this section, we compare our proposed approach with other relevant studies. It is worth noting that most of the existing literature on automatic polyp detection employs varying experimental settings and evaluation criteria. Furthermore, they often use different datasets for training and testing purposes. Despite these discrepancies, we have opted to benchmark our method on the SUN dataset, which is the largest and most recent polyp dataset. The experimental results are presented in Table 6.

Misawa et al. developed the SUN dataset and have been consistently updating their latest models' performance on the dataset's website (http://amed8k.sundatabase.org). The accuracy of the latest approved model (July 2022) is shown in Table 6 No. 1. The authors presented their results in a confusion matrix and reported the sensitivity and the specificity. To facilitate comparison with other studies, we calculated metrics those were not specifically mentioned in the original paper, which are denoted in the table with *. Notably, the highest precision was observed in their results. Puyal et al. (No. 2) acknowledged that polyp detection algorithms are commonly utilized in colonoscopy videos. Consequently, they developed a hybrid 2D/3D CNN that integrates temporal information to identify polyps in video data. Ishak Pacal et al. (No. 3) improved the YOLOv3 and YOLOv4 algorithms in several aspects, including the backbone network, the activation function, the loss function, and the training strategy. Karaman et al. (No. 4 & 5) proposed an effective hyperparameter optimization method called "Artificial Bee Colony" (ABC), which can be easily applied to various versions of the YOLO algorithm, including YOLOv3, YOLOv4, Scaled-YOLOv4, YOLOv5, YOLOR, and YOLOv7. Finally, they discussed the performance of their proposed approach on the task of polyp detection.

Overall, our model shows successful performance across all performance metrics, particularly in recall where it significantly outperforms the current state-of-the-art results. Additionally, our model ranks second in precision and mAP, with a difference of 0.87% and 1.02%, respectively, compared to the best results.

Table 6
Comparison of the performance between the proposed model and the state-of-the-art models. The values marked with * are not reported in the author's article. We use the F1 calculation formula to derive them.

| No. | Research Group | Method | dataset | precision | recall | F1 | mAP |
|---|---|---|---|---|---|---|---|
| 1 | Masashi Misawa et al., 2022 [17] | | SUN | **99.65*** | 91.50 | 95.40* | NA |
| 2 | Juana González-Bueno Puyal. et al., 2022 [53] | hybrid 2D/3D CNN | SUN | 88.18* | 86.99 | 87.58 | NA |
| 3 | Ishak Pacal et al., [38] | Proposed YOLOv3 | SUN | 96.49 | 86.48 | 91.21 | **99.21** |
| 4 | Ahmet Karaman et al., [27] | YOLOv4-CSP+ABC | SUN | 86.00 | 84.00 | 85.00 | 89.00 |
| 5 | Ahmet Karaman et al., 2023 [39] | YOLOv5x + ABC | SUN | 93.44 | 81.35 | 86.98* | 92.17 |
| 6 | Ours | YOLO-OB | SUN | 98.37 | **98.23** | **98.32** | 98.19 |

4.6 Limitations and future directions

The proposed model has exhibited successful performance in terms of detection accuracy on the publicly available dataset SUN, as well as our private dataset Union. In this study, we only used single-frame images as experimental data. However, polyps appear continuously over time in real colonoscopy videos. Continuous frames can help avoid common problems in static image object detection, such as blurry or glare frames. In the future, we plan to introduce a time dimension into the algorithm to improve the accuracy of polyp tracking using colonoscopy videos. During our efforts to promote polyp tracking task, we encountered low confidence issues in YOLO-OB's prediction results. Although this issue may not significantly impact the detection results, it can still affect the tracking program's ability to distinguish false positive detection boxes. How to further improve the confidence of predicted bounding boxes for polyp detection algorithm, and thus promote the development of video-based polyp tracking, is our future research direction.

**5. Conclusion**

Based on the basic framework of the YOLO series algorithms, this paper proposes a polyp detection model called YOLO-OB. Compared with the current state-of-the-art models, our model can effectively improve the recall rate for polyp detection. We focused on multi-scale detection problem and made improvements in several aspects: data augmentation, multi-scale feature fusion, and detection head module. To verify the generalization performance of the model, we manually labeled the polyp dataset Union, which was collected from different hospitals, to further expand the training and testing samples.

For the data preprocessing, we manually constructed more diverse-sized training samples using mosaic data augmentation. For the feature fusion, we followed the multi-resolution detection approach and used feature maps of different depths in DarkNet53 to achieve multi-scale polyp detection. We also designed a BiSPFPN layer to efficiently fuse feature maps at different depths of the backbone network. To improve the performance of detection head, we designed an anchor-free box regression method based on ObjectBox.

Our proposed model achieved high performance on both the public dataset SUN and the private dataset Union. The YOLO-OB can achieve real-time polyp detection at 39 FPS on a RTX3090, and 29 FPS on a GTX1080Ti graphic card. Therefore, it can be implemented in a real time colonoscopy CAD system to aid physicians in detecting colon polyps. The implementation of this work can decrease the occurrence of missed polyps and consequently lower the risk of patients developing colon cancer.

**Funding**

This work was supported by the Wuhan Major Science and Technology Project for Breaking Through the Key Bottleneck Technologies (No. 2021022002023426).

**Contributors**

Xiao Yang: Methodology, Data analysis and interpretation, Manuscript preparation, Manuscript editing.

Guangzhi Ma: Quality control of data and algorithms, Manuscript editing, Manuscript review.

Enmin Song: Study concepts, Manuscript review.

Yunfeng Zhu: Study design, Experimental design, Acquisition of data.

Dongming Yu: Study design, Acquisition of data.

Bowen Ding: Study design, Acquisition of data.

Xianyuan Wang: Colonoscopy dataset support, Polyp ground truth label validation.

**Research data for this article**

The Python implementation of this paper can be found here:
https://doi.org/10.17632/xj3x7vmsdg.1
https://github.com/seanyan62/YOLO-OB

**Declaration of competing interest**

The authors declare that there is no conflict of interest related to this paper.

**Reference**

[1] R.L. Siegel, N.S. Wagle, A. Cercek, R.A. Smith, A. Jemal, Colorectal cancer statistics, 2023, CA: a cancer journal for clinicians, 73 (2023) 233-254.https://doi.org/10.3322/caac.21772

[2] R.L. Siegel, K.D. Miller, N.S. Wagle, A. Jemal, Cancer statistics, 2023, Ca-Cancer J Clin, 73 (2023) 17-48.https://doi.org/10.3322/caac.21763

[3] I.A. Issa, M. Noureddine, Colorectal cancer screening: An updated review of the available options, World journal of gastroenterology, 23 (2017) 5086.https://doi.org/10.3748/wjg.v23.i28.5086

[4] G.P. Ji, G.B. Xiao, Y.C. Chou, D.P. Fan, K. Zhao, G. Chen, L. Van Gool, Video Polyp Segmentation: A Deep Learning Perspective, Mach Intell Res, 19 (2022) 531-549.https://doi.org/10.1007/s11633-022-1371-y

[5] P. Wang, P.X. Liu, J.R.G. Brown, T.M. Berzin, G.Y. Zhou, S. Lei, X.G. Liu, L.P. Li, X. Xiao, Lower Adenoma Miss Rate of Computer-Aided Detection-Assisted Colonoscopy vs Routine White-Light Colonoscopy in a Prospective Tandem Study, Gastroenterology, 159 (2020) 1252-+.https://doi.org/10.1053/j.gastro.2020.06.023

[6] G. Urban, P. Tripathi, T. Alkayali, M. Mittal, F. Jalali, W. Karnes, P. Baldi, Deep learning localizes and identifies polyps in real time with 96% accuracy in screening colonoscopy, Gastroenterology, 155 (2018) 1069-1078. e1068.https://doi.org/10.1053/j.gastro.2018.06.037

[7] K. ELKarazle, V. Raman, P. Then, C. Chua, Detection of Colorectal Polyps from Colonoscopy Using Machine Learning: A Survey on Modern Techniques, Sensors, 23 (2023) 1225.https://doi.org/10.3390/s23031225

[8] I. Barua, D.G. Vinsard, H.C. Jodal, M. Løberg, M. Kalager, Ø. Holme, M. Misawa, M. Bretthauer, Y. Mori, Artificial intelligence for polyp detection during colonoscopy: a systematic review and meta-analysis, Endoscopy, 53 (2021) 277-284.https://doi.org/10.1055/a-1201-7165

[9] N. Hoerter, S.A. Gross, P.S. Liang, Artificial intelligence and polyp detection, Current treatment options in gastroenterology, 18 (2020) 120-136.https://doi.org/10.1007/s11938-020-00274-2

[10] Z. Zou, K. Chen, Z. Shi, Y. Guo, J. Ye, Object detection in 20 years: A survey, Proceedings of the IEEE, (2023).https://doi.org/10.1109/jproc.2023.3238524

[11] S. Ameling, S. Wirth, D. Paulus, G. Lacey, F. Vilarino, Texture-Based Polyp Detection in Colonoscopy, Springer Berlin Heidelberg, Berlin, Heidelberg, 2009, pp. 346-350.https://doi.org/10.1007/978-3-540-93860-6_70

[12] S.Y. Park, D. Sargent, I. Spofford, K.G. Vosburgh, Y. A-Rahim, A Colon Video Analysis Framework for Polyp Detection, IEEE T Bio-Med Eng, 59 (2012) 1408-1418.https://doi.org/10.1109/Tbme.2012.2188397

[13] S.A. Karkanis, D.K. Iakovidis, D.E. Maroulis, D.A. Karras, M. Tzivras, Computer-aided tumor detection in endoscopic video using color wavelet features, IEEE T Inf Technol B, 7 (2003) 141-152.https://doi.org/10.1109/Titb.2003.813794

[14] D.E. Maroulis, D.K. Iakovidis, S.A. Karkanis, D.A. Karras, CoLD: a versatile detection system for colorectal lesions in endoscopy video-frames, Comput Meth Prog Bio, 70 (2003) 151-166.https://doi.org/10.1016/s0169-2607(02)00007-x

[15] D.K. Iakovidis, D.E. Maroulis, S.A. Karkanis, An intelligent system for automatic detection of gastrointestinal adenomas in video endoscopy, Comput Biol Med, 36 (2006) 1084-1103.https://doi.org/10.1016/j.compbiomed.2005.09.008

[16] D.A. Corley, C.D. Jensen, A.R. Marks, W.K. Zhao, J.K. Lee, C.A. Doubeni, A.G. Zauber, J. de Boer, B.H. Fireman, J.E. Schottinger, V.P. Quinn, N.R. Ghai, T.R. Levin, C.P. Quesenberry, Adenoma Detection Rate and Risk of Colorectal Cancer and Death, New Engl J Med, 370 (2014) 1298-1306.https://doi.org/10.1056/NEJMoa1309086

[17] M. Misawa, S.E. Kudo, Y. Mori, K. Hotta, K. Ohtsuka, T. Matsuda, S. Saito, T. Kudo, T. Baba,

F. Ishida, H. Itoh, M. Oda, K. Mori, Development of a computer-aided detection system for colonoscopy and a publicly accessible large colonoscopy video database (with video), Gastrointest Endosc, 93 (2021) 960-+.https://doi.org/10.1016/j.gie.2020.07.060

[18] E. Elizar, M.A. Zulkifley, R. Muharar, M.H.M. Zaman, S.M. Mustaza, A Review on Multiscale-Deep-Learning Applications, Sensors, 22 (2022).https://doi.org/10.3390/s22197384

[19] S. Lafraxo, M. Souaidi, Z. Kerkaou, M. El Ansari, L. Koutti, A Multiscale Polyp Detection Approach for GI Tract Images Based on Improved DenseNet and Single-Shot Multibox Detector, Diagnostics, 13 (2023).https://doi.org/10.3390/diagnostics13040733

[20] S. Zhang, J.R. Wu, E.Z. Shi, S.G. Yu, Y.F. Gao, L.C. Li, L.R. Kuo, M.J. Pomeroy, Z.J. Liang, MM-GLCM-CNN: A multi-scale and multi-level based GLCM-CNN for polyp classification, Comput Med Imag Grap, 108 (2023).https://doi.org/10.1016/j.compmedimag.2023.102257

[21] C. Cao, R. Wang, Y. Yu, H. Zhang, Y. Yu, C. Sun, Gastric polyp detection in gastroscopic images using deep neural network, PloS one, 16 (2021) e0250632.https://doi.org/10.1371/journal.pone.0250632

[22] S. Wang, Y. Cong, H. Zhu, X. Chen, L. Qu, H. Fan, Q. Zhang, M. Liu, Multi-scale context-guided deep network for automated lesion segmentation with endoscopy images of gastrointestinal tract, IEEE Journal of Biomedical and Health Informatics, 25 (2020) 514-525.https://doi.org/10.1109/jbhi.2020.2997760

[23] Y. Wang, W. Tavanapong, J. Wong, J.H. Oh, P.C. de Groen, Polyp-Alert: Near real-time feedback during colonoscopy, Comput Meth Prog Bio, 120 (2015) 164-179.https://doi.org/10.1016/j.cmpb.2015.04.002

[24] B.B.S.L. Houwen, K.J. Nass, J.L.A. Vleugels, P. Fockens, Y. Hazewinkel, E. Dekker, Comprehensive review of publicly available colonoscopic imaging databases for artificial intelligence research: availability, accessibility, and usability, Gastrointest Endosc, 97 (2023).https://doi.org/10.1016/j.gie.2022.08.043

[25] I. Pacal, D. Karaboga, A. Basturk, B. Akay, U. Nalbantoglu, A comprehensive review of deep learning in colon cancer, Comput Biol Med, 126 (2020).https://doi.org/10.1016/j.compbiomed.2020.104003

[26] J. Bernal, N. Tajkbaksh, F.J. Sánchez, B.J. Matuszewski, H. Chen, L.Q. Yu, Q. Angermann, O. Romain, B. Rustad, I. Balasingham, K. Pogorelov, S. Choi, Q. Debard, L. Maier-Hein, S. Speidel, D. Stoyanov, P. Brandao, H. Cordova, C. Sánchez-Montes, S.R. Gurudu, G. Fernández-Esparrach, X. Dray, J.M. Liang, A. Histace, Comparative Validation of Polyp Detection Methods in Video Colonoscopy: Results From the MICCAI 2015 Endoscopic Vision Challenge, IEEE T Med Imaging, 36 (2017) 1231-1249.https://doi.org/10.1109/Tmi.2017.2664042

[27] A. Karaman, D. Karaboga, I. Pacal, B. Akay, A. Basturk, U. Nalbantoglu, S. Coskun, O. Sahin, Hyper-parameter optimization of deep learning architectures using artificial bee colony (ABC) algorithm for high performance real-time automatic colorectal cancer (CRC) polyp detection, Appl Intell, 53 (2023) 15603-15620.https://doi.org/10.1007/s10489-022-04299-1

[28] Z.-Q. Zhao, P. Zheng, S.-t. Xu, X. Wu, Object detection with deep learning: A review, IEEE transactions on neural networks and learning systems, 30 (2019) 3212-3232.https://doi.org/10.1109/TNNLS.2018.2876865

[29] X. Mo, K. Tao, Q. Wang, G. Wang, An efficient approach for polyps detection in endoscopic videos based on faster R-CNN,  2018 24th international conference on pattern recognition (ICPR), IEEE, 2018, pp. 3929-3934.https://doi.org/10.1109/icpr.2018.8545174

[30] R. Girshick, Fast r-cnn,  Proceedings of the IEEE international conference on computer vision, 2015, pp. 1440-1448.https://doi.org/10.1109/iccv.2015.169

[31] T.Y. Lin, P. Dollar, R. Girshick, K.M. He, B. Hariharan, S. Belongie, Feature Pyramid Networks for Object Detection, 30th IEEE Conference on Computer Vision and Pattern Recognition (Cvpr 2017), (2017) 936-944.https://doi.org/10.1109/Cvpr.2017.106

[32] M. Murugesan, R.M. Arieth, S. Balraj, R. Nirmala, Colon cancer stage detection in colonoscopy images using YOLOv3 MSF deep learning architecture, Biomed Signal Proces, 80 (2023).https://doi.org/10.1016/j.bspc.2022.104283

[33] S. Liu, L. Qi, H.F. Qin, J.P. Shi, J.Y. Jia, Path Aggregation Network for Instance Segmentation, Proc Cvpr IEEE, (2018) 8759-8768.https://doi.org/10.1109/Cvpr.2018.00913

[34] T.-Y. Lin, M. Maire, S. Belongie, J. Hays, P. Perona, D. Ramanan, P. Dollár, C.L. Zitnick, Microsoft coco: Common objects in context,  Computer Vision–ECCV 2014: 13th European Conference, Zurich, Switzerland, September 6-12, 2014, Proceedings, Part V 13, Springer, 2014, pp. 740-755.https://doi.org/10.1007/978-3-319-10602-1_48

[35] G. Falcao, P. Carrinho, Highly accurate and fast YOLOv4-based polyp detection, Expert Systems with Applications, (2023) 120834.https://doi.org/10.1016/j.eswa.2023.120834

[36] I. Pacal, D. Karaboga, A robust real-time deep learning based automatic polyp detection system, Comput Biol Med, 134 (2021).https://doi.org/10.1016/j.compbiomed.2021.104519

[37] M. Tan, R. Pang, Q.V. Le, EfficientDet: Scalable and Efficient Object Detection,  2020 IEEE/CVF Conference on Computer Vision and Pattern Recognition (CVPR), 2020, pp. 10778-10787.https://doi.org/10.1109/cvpr42600.2020.01079

[38] I. Pacal, A. Karaman, D. Karaboga, B. Akay, A. Basturk, U. Nalbantoglu, S. Coskun, An efficient real-time colonic polyp detection with YOLO algorithms trained by using negative samples and large datasets, Comput Biol Med, 141 (2022).https://doi.org/10.1016/j.compbiomed.2021.105031

[39] A. Karaman, I. Pacal, A. Basturk, B. Akay, U. Nalbantoglu, S. Coskun, O. Sahin, D. Karaboga, Robust real-time polyp detection system design based on YOLO algorithms by optimizing activation functions and hyper-parameters with artificial bee colony (ABC), Expert Systems with Applications, 221 (2023).https://doi.org/10.1016/j.eswa.2023.119741

[40] A. Nogueira-Rodriguez, R. Dominguez-Carbajales, F. Campos-Tato, J. Herrero, M. Puga, D. Remedios, L. Rivas, E. Sanchez, A. Iglesias, J. Cubiella, F. Fdez-Riverola, H. Lopez-Fernandez, M. Reboiro-Jato, D. Glez-Pena, Real-time polyp detection model using convolutional neural networks, Neural Comput Appl, 34 (2022) 10375-10396.https://doi.org/10.1007/s00521-021-06496-4

[41] J. Redmon, A. Farhadi, Yolov3: An incremental improvement, arXiv preprint arXiv:1804.02767, (2018).https://doi.org/10.48550/arXiv.1804.02767

[42] C.-Y. Wang, A. Bochkovskiy, H.-Y.M. Liao, YOLOv7: Trainable bag-of-freebies sets new state-of-the-art for real-time object detectors,  Proceedings of the IEEE/CVF Conference on Computer Vision and Pattern Recognition, 2023, pp. 7464-7475.https://doi.org/10.48550/arXiv.2207.02696

[43] P. Viola, M. Jones, Rapid object detection using a boosted cascade of simple features, Proceedings of the 2001 IEEE computer society conference on computer vision and pattern recognition. CVPR 2001, IEEE, 2001, pp. I-I.https://doi.org/10.1109/cvpr.2001.990517

[44] N. Dalal, B. Triggs, Histograms of oriented gradients for human detection,  2005 IEEE computer society conference on computer vision and pattern recognition (CVPR'05), IEEE, 2005, pp. 886-893.https://doi.org/10.1109/cvpr.2005.177

[45] S.Q. Ren, K.M. He, R. Girshick, J. Sun, Faster R-CNN: Towards Real-Time Object Detection

with Region Proposal Networks, Advances in Neural Information Processing Systems 28 (Nips 2015), 28 (2015).https://proceedings.neurips.cc/paper_files/paper/2015/hash/14bfa6bb14875e45bba028a21ed38046-Abstract.html

[46] R.K. Zhang, Y.L. Zheng, C.C.Y. Poon, D.G. Shen, J.Y.W. Lau, Polyp detection during colonoscopy using a regression-based convolutional neural network with a tracker, Pattern Recogn, 83 (2018) 209-219.https://doi.org/10.1016/j.patcog.2018.05.026

[47] D.C. Wang, N. Zhang, X.Z. Sun, P.F. Zhang, C.X. Zhang, Y. Cao, B.Y. Liu, AFP-Net: Realtime Anchor-Free Polyp Detection in Colonoscopy, Proc Int C Tools Art, (2019) 636-643.https://doi.org/10.1109/Ictai.2019.00094

[48] X. Sun, D. Wang, Q. Chen, J. Ni, S. Chen, X. Liu, Y. Cao, B. Liu, MAF-Net: Multi-branch anchor-free detector for polyp localization and classification in colonoscopy, International Conference on Medical Imaging with Deep Learning, PMLR, 2022, pp. 1162-1172.https://proceedings.mlr.press/v172/sun22a.html

[49] K.W. Duan, S. Bai, L.X. Xie, H.G. Qi, Q.M. Huang, Q. Tian, CenterNet: Keypoint Triplets for Object Detection, IEEE I Conf Comp Vis, (2019) 6568-6577.https://doi.org/10.1109/Iccv.2019.00667

[50] M. Zand, A. Etemad, M. Greenspan, ObjectBox: From Centers to Boxes for Anchor-Free Object Detection, Computer Vision, Eccv 2022, Pt X, 13670 (2022) 390-406.https://doi.org/10.1007/978-3-031-20080-9_23

[51] S. Elfwing, E. Uchibe, K. Doya, Sigmoid-weighted linear units for neural network function approximation in reinforcement learning, Neural Networks, 107 (2018) 3-11.https://doi.org/10.1016/j.neunet.2017.12.012

[52] O. Ronneberger, P. Fischer, T. Brox, U-Net: Convolutional Networks for Biomedical Image Segmentation, Medical Image Computing and Computer-Assisted Intervention, Pt Iii, 9351 (2015) 234-241.https://doi.org/10.1007/978-3-319-24574-4_28

[53] J. Gonzalez-Bueno Puyal, P. Brandao, O.F. Ahmad, K.K. Bhatia, D. Toth, R. Kader, L. Lovat, P. Mountney, D. Stoyanov, Polyp detection on video colonoscopy using a hybrid 2D/3D CNN, Med Image Anal, 82 (2022) 102625.https://doi.org/10.1016/j.media.2022.102625